\documentclass[10pt, a4,twocolumn]{article}
\usepackage{cite}
\usepackage{amsmath,amssymb,amsfonts}
\usepackage{algorithmic}
\usepackage{graphicx}
\usepackage{textcomp}
\usepackage{xcolor}
\def\BibTeX{{\rm B\kern-.05em{\sc i\kern-.025em b}\kern-.08em
    T\kern-.1667em\lower.7ex\hbox{E}\kern-.125emX}}
\newtheorem{defi}{Definition}{\bf}{\it}
{\bf}{\it}
    
\begin{document}
\graphicspath{{PICTS/}}

\title{A novel HD Computing Algebra: Non-associative superposition of states creating sparse bundles representing order information}
\author{Stefan Reimann \\
              Institute of Neuroinformatics\\
	University of Zurich and ETH Zurich\\
              Switzerland\\
}
\maketitle
\begin{abstract}
Information inflow into a computational system is by a sequence of information items. Cognitive computing, i.e. performing transformations along that sequence, requires to represent item information as well as sequential information. Among the most elementary operations is bundling, i.e. adding items, leading to 'memory states', i.e. bundles, from which information can be retrieved. If the bundling operation used is associative, e.g. ordinary vector-addition, sequential information can not be represented without imposing additional algebraic structure. A simple stochastic binary bundling rule inspired by the stochastic summation of neuronal activities allows the resulting memory state to represent both, item information as well as sequential information as long as it is non-associative. The memory state resulting from bundling together an arbitrary number of items is non-homogeneous and has a degree of sparseness, which is controlled by the activation threshold in summation. The bundling operation proposed allows to build a filter in the temporal as well as in the items' domain, which can be used to navigate the continuous inflow of information.
\end{abstract}


\section{Introduction}
Cognitive agents are subject to a continuous inflow of sensory information, which is permanently processed to direct behavioural acts. Such a sequence contains information about the single items, about potential associations between items, as well as information about their sequential order \cite{murdock1974human}. This is evident when considering an animal moving through a landscape: the input flow is a sequence of, among others, visual or olfactory stimuli; reading a text provides one with a sequence of information items, i.e. words. Items may  be correlated or 'similar' to each other. This similarity can reflect continuity, when moving on a continuous path, or dependences between items, e.g. words in a meaningful sentence. 

Typically the degree of similarity of two items is measured by the so-called cosine similarity, meaning that similarity is related to the goodness of how well one item can be projected onto the other. In a broader sense, similarity is related to distance, requiring that the 'space of items' admits a 'metric'. When representing items by binary sequences, the natural space of items is the Hamming space with the Hamming distance being the metric defined on it. Other choices are possible, e.g. the Jaccard distance or the edit distance, depending on particular requirements or interests. Additionally, many cognitive tasks require to also discriminate between items. B. Murdock writes \cite{murdock1993todam2}: "{\it The recall probability depends on two factors. The first factor is how close [for two states $g$ and $g'$] the dot product $ g' \cdot g$ is to .. [ $1$, identity]. The second factor is whether any other item (e.g.$h$) is closer to $1$ than $g'$ is.}". The cosine similarity of two states may be $\frac{1}{4}$, but this value does not say anything about how distinguishable the two items actually are in item space. Two visual figures may be better discriminated from each other, the sharper the picture is or the more colours are available to represent them, i.e. the higher the resolution is. In the sense of Murdock, we define the distance of two items as a function of the probability to find another object 'in between them', i.e. having smaller distance to either of these than the two items have \cite{reimann2021computing}, see eq. \ref{eq:D}. This distance measure incorporates global information about the state space. The recall probability might be defined as some increasing function of that distance. For example, if the dimension of the state space is small, the states are close, while if it is huge, states are actually quite far away from each other.

 Sequence learning as well as discrimination are important topics in computer science as well as in cognitive sciences. In the following, we will consider a particular computational architecture, whose fundamental framework is: "A physical item evokes a neural activity pattern, represented by a high dimensional random vector, called a state-vector. Computation is by a sequence of transformation applied to this pattern. Transformations are assumed to be elementary addition for bundling and multiplication for binding."

Corresponding architectures, also called 'Vector Symbolic Architectures' (VSA's), thus are particular algebra's for High-Dimensional Computing, i.e. a set of states together with a collection of arithmetic operations determining the rules of computation on that state-space, see \cite{schlegel2020comparison} for a comparison of different High-Dimensional algebras (HD-algebras). The model for "storage and retrieval of item and associative information" proposed by B. Murdock in 1982 \cite{murdock1982theory} is a full-blown VSA-model in today's terms: {\it Items are represented by (continuous) random vectors, associations are represented by the convolution of item vectors, information is stored in a common memory vector. Retrieval is by correlation, the approximate inverse of convolution.}

The algebra considered in the following was first proposed in \cite{reimann2021algebra} in the context of human working memory. Therein it is shown that experimentally observed data concerning different cognitive tasks can be well described by that elementary algebra. 

\begin{figure}[h]
{\center
		\includegraphics[width=2.2in]{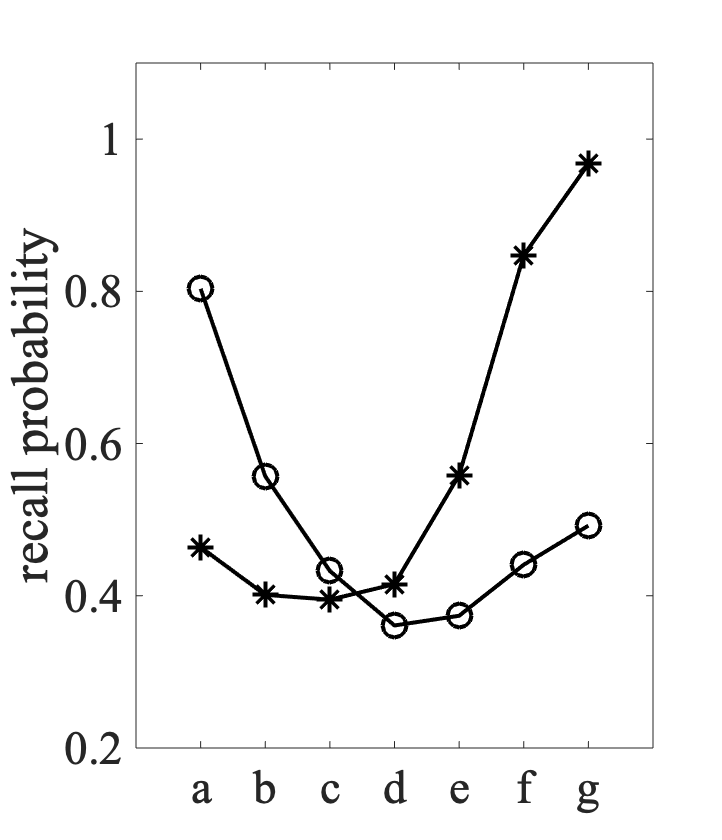}
		\caption{The Serial Position Curves simulated in the context of the cognitive algebra proposed for items $a \hdots g$, see \cite{reimann2021algebra}. The Serial Position Curve displays the typical $U$-shape, i.e. prominent primacy or recency effects, as well known from experiments. For a brief overview about typical experimental findings see \cite{lewandowsky1989memory}}\label{fig:Curves}
		}		
\end{figure}
\subsection{Bundling: associative vs. non-associative}
HD-algebras share some fundamental motives: States are high-dimensional random vectors drawn from some distribution over some field, their similarity is quantified by a normalised dot-product, while their superposition is by vector-addition, exceptions being the classes belonging to Kanerva's Binary Spatter Code \cite{kanerva2009hyperdimensional}. 

A fundamental property of 'vector-addition' is that it is associative, i.e. for any three states $x+(y+z) = (x+y) + z$, so that the sum is independent of the order of components. By so bundling the sequence $(x,y,z)$, information about order, i.e. sequential information, is lost. In tasks in which ordering is essential, vector-addition therefore appears not to be adequate. To recover the sequence from the sum, the ordering has to be separately coded or otherwise re-established. Additional computational structure is needed. Various possibilities exist. Given a list of states $a,b,c$, that list of single states can be enriched in various ways, which are described below:

\begin{equation*}\label{eq:cognRep}
\begin{pmatrix}
a\\b\\c
\end{pmatrix} \quad \longmapsto \quad
\begin{pmatrix} a&\rho^2\\b&\rho^1\\c&\rho^0\end{pmatrix} \; , \;
\begin{pmatrix} a&1\\b&2\\c&3\end{pmatrix} \; , \;
\begin{pmatrix} a&\eta\\b&a\\c&b\end{pmatrix} .
\end{equation*}
where $\eta$ is some pre-experimental or initial state. 
\begin{enumerate}
\item Extend the set of algebraic rules, i.e. addition and multiplication, by a third one, $\rho$, which maps a state $x$ to its permutation $\rho\:x$. The sequence $(x,y)$ is then mapped on the state
$
{\bf M}:= \rho * x + y.
$
Sequences are coded in terms of polynomials in $\rho$, see \cite{kanerva2009hyperdimensional}. Serial order information is in the powers of the permutation operator $\rho$. Note that in order to retrieve $x$ from $\bf M$, the inverse of addition as well as the inverse $\rho$ must be known, i.e. $x = \rho^{-1}({\bf M}-y)$. 

\item Coding the sequence in absolute terms requires to extent the state $a$ corresponding to an item $A$ by a serial-position marker $t_a$, e.g. $(a,1)$ meaning that item $A$ occurs at the first position in the list. This possibility is of course closely related to coding ordering by a decreasing number of powers of permutations.
\item Serial position can also be encoded in relative terms in that each state is bound to its precursor. The sequence $(a,b)$ corresponding to items $(A,B)$ is represented by the sum ${\bf M} =a*\eta + b*a$, where $\eta$ is some initial or pre-experimental state, while the second term $b*a$ means that $b$ is bound to its neighbour $a$. This mechanism is related to chaining in memory models. 

\item Order may be coded by postulating an activation gradient \cite{page1998primacy} as used in many models of human WM: subsequent states are equipped with different activations, e.g. the sequence $(x,y)$ is attached to a real-valued activation gradient ${\cal A}$, so that ${\cal A}(x) < {\cal A}(y)$. Together with postulating that a state with higher activation can be more easily retrieved, this gives rise to an order effect, i.e. more recent items are better retrieved than earlier ones \cite{page1998primacy}. If the activation gradient is reversed, i.e. ${\cal A}(x) > {\cal A}(y)$, this postulate results in a recency effect. This strategy also appears in TODAM, as Murdock parametrises weights according to their order \cite{lewandowsky1989memory}. 
\end{enumerate}

Without question, associativity as resulting from bundling by vector-addition is computationally very convenient. But the price to be paid is the loss of order information and the effort of additional algebraic structure to re-establish serial information. What, if one allows non-associative bundling, without adding structure to the algebra? Many operations are not associative, including fundamental ones such as subtraction and division, exponentiation, or the assignment and conditional operators in programming. Convex summation, defined by $x +_c y := cx+(1-c)y$ for some constant $0 < c < 1$ is not associative unless $c = \frac{1}{2}$. An other example is taken from Etherington \cite{etherington1941ii} and concerns mating of species: Let $x,y,z$ denote different species and let $\alpha(x,y)$ be the offspring of $x$ mating with $y$. Then, according to the Mendelian Laws of genetics, the offspring of $z$ mating the offspring of $x$ and $y$ is (in general) different from the offspring of $x$, mating the offspring of $y$ and $z$, i.e. mating is a not associative operation 
\begin{equation}
\alpha\big(\alpha(x,y),z\big) \not= \alpha\big(x,\alpha(y,z)\big).
\end{equation}

In case of non-associativity, a unique sequence of states gives rise to two different bundles: the $\bf L$-bundles results from left-associative operating and the $\bf R$-bundle results from right-associative operating, which are different, see Sec. \ref{sec:bundles}. 
As shown below, both bundles code inverse sequential information and become quasi-orthogonal if the length of the memory list increases. 
In that sense, it is non-associativity which enriches presentation, but not by a proposed additional structure but as a consequence of the bundling operation. As both bundles are related to the same sequence of items, they might be regarded as two components of the corresponding memory state
\begin{equation}\label{eq:M}
{\bf M} =\begin{pmatrix} {\bf L }\\ {\bf R}\end{pmatrix}.
\end{equation}

The rest of the paper is organised as follows: In section \ref{sec:algebra} we define the algebra of patterns to work with. It brings together three biologically plausible and important concepts: high-dimensionality, randomness, and sparseness. Binding is by correlation, while bundling mimics the stochastic addition of neuronal activity and it shown to be non-associative. Bundling a sequence of items in a memory list, as described in Sec \ref{sec:bundles}, creates two memory states representing that sequence, i.e. item as well as order information, see Sec \ref{seq:distances}. Memory states are shown to be informative for arbitrary list length, i.e. memory vector never becomes homogeneous, if bundling is non-associative. The bundling operation defined creates sparseness, as shown in \ref{sec:sparseness}. The parameter governing the bundling operation allows to fine tune the sparseness of the memory states.

The regulation of sparseness gains its importance from the fact that (metabolic) energy is limited, while neuronal computation is costly. Sparseness allows to balance representational capacity with memory capacity. Maximising representational capacity as well as memory capacity, given limited energy, requires sparse pattern. Corresponding fine-tuning can be realised in the bundling operation $+_p$ by changing the threshold $p$.

The set of all possible activity patterns of length $N$ is called the state-space $\mathbb{X}^N$ of the computational system, and its elements representing activity patterns are called states. The representational capacity, i.e. the number of different states of length $N$, is $2^N$, while there are ${N \choose qN}$ states of length $N$ with mean activity $q$. This number is maximal for $q  \frac{1}{2}$ and increases exponentially in $N$. These dense states are costly in terms of energy. Due to this energy constraint, high dimensional representation is only possible for sufficiently sparse states. On the other hand, the less dense states are, the lower their interference, i.e. memory capacity increases. Sparseness allows to balance representational capacity with memory capacity.

\section{The algebra of cognitive states $\left( \mathbb{X}, \oplus_p, \ast\right)$}\footnote{We use $\oplus_p$ for addition of states to avoid confusion with commly addition $+$ over real numbers.}\label{sec:algebra}
Cognitive function are based on computations which are implemented in a huge network of neuronal components. This statement implies three questions: 1.) What are the elements computed with? What are the rules according to which computation is performed? How are physical items represented by these elements? 
The first two questions concern the structure of an appropriate algebraic structure, while the third one is how an item is represented as an element of algebra.

\subsection{The state-space}  In the course of perception, a physical item induces a sensory input, which evokes an activity pattern in the neuronal field it is projected to. This activity pattern is a transformation of the activity pattern corresponding to the sensory input. That way, a physical item is represented by a binary pattern, in which $1's$ indicate active neurons, while $0's$ indicate inactive ones. Due to the size and structural complexity of the neuronal correlate, patterns are described by high-dimensional random binary vectors. These binary vectors are called {\it states}. A state thus is an activation pattern representing a physical item. The state space is the set of all possible representations. Component-wise operations such as $AND., OR.$ and $XNOR.$ are generically defined on the state-space and thus can be used to define further operations on it. 

It is assumed that there is a metric defined on the set of states. For example, the normalised Hamming distance between two states $x$ and $y$ yields 
\begin{equation}
d(x,y) := \frac {1}{N} \: |XOR.(x,y)|,
\end{equation}
where $|x| = \sum_i x_i$ is the 1-norm of the binary vector $x$.

Other metrics can be defined on the set of states, being adapted to particular needs and assumptions, e.g. the Jaccard distance, which may serve as an intuitive measure for sparse states. 
Note that the Jaccard distance of two states having average activity $q$, $0<q\leq 1$, is proportional to their normalised Hamming distance given by $d(x,y) = q(2-q) \: d_J(x,y)$.

This distance between two states is local and does not say anything about the closeness of the two states in the state space. Two states are called close if they are hard to distinguish. As a measure for the closeness of two states we take the probability to find another state by chance which is closer to $x$ than $y$. This concerns the distribution function of distances on the set of states, which provides the global information about the set of states, which will enter the distance. The global distance $D$ derived from the local distance $d$ is defined as 
\begin{equation}\label{eq:D}
D(x,y) := \mathbb{P}_{\mathbb{X}^N} \Bigl[\tilde{d}\leq d(x,y)\Bigr], \;0\leq \tilde{d}\leq 1
\end{equation} 
Note that $D(x,y)$ is decreasing in $N$, while it is increasing in $d = d(x,y)$. 

\begin{defi}
Two states are equal to each other, $x \approx y$, if they are sufficiently close to each other, i.e. $D(x,y) \leq \epsilon$, for some small non-negative $\epsilon$.
\end{defi}
Note that two states can be equal in high dimensions, while in lower dimension they are not. On the other hand, equals states will remain equal if dimension is increased. 
\footnote{
If two states $x$ and $y$ are equal in $\mathbb{X}^N$, then they are also similar in $\mathbb{X}^{\hat{N}}$, where $\hat{N}> N$.}
Summing up, the state space is defined as Hamming space with global metric $D$.

\begin{defi}[The state space]
The state space $\mathbb{X}^N$ is the set of all binary sequences of length $N$ equipped with metric $D$, defined as in eq \ref{eq:D}. The subset of all states with mean activity $q$ is denoted by $\mathbb{X}^N_q$. 
\end{defi} 
The notion "$x$ and $y$ are states!" means that $x,y \in \mathbb{X}^N$, while a state with mean activity $q$ will be called a 'q-state'. 

\subsection{The operations on the state-space: addition and multiplication}
Two operations will be defined on the state space. They correspond to binding and bundling. Two items are (associatively) bond to each other, if one can be retrieved by cueing with the other item. Bundling refers to collecting items by superimposing their respective states. The corresponding formal operations are {\it multiplication} and {\it addition}, defined in the following. 

\begin{figure}[h]
{\center
\includegraphics[width=2.6in]{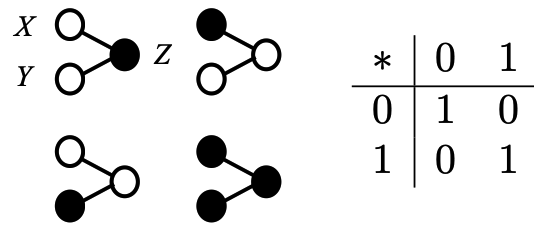} 
\caption{Binding patterns by coincidence: Neuron Z is active whenever neurons X and Y are in the same state.}\label{fig:multiplication}
}
\end{figure}

\paragraph{\bf Binding by multiplication}
Binding is proposed to be by coincidence, i.e. two items are the stronger bound to each other the more simultaneously activated components are in the respective neural patterns. The corresponding formal operation is component-wise $XNOR.$ as shown in Fig \ref{fig:multiplication}.

Its is worth noting that there is an immediate relation between Hamming distance and multiplication:
\begin{equation}\label{eq:DistanceMultiplication}
d(x,y) = 1 - \frac{|x*y|}{N}.
\end{equation}
This allows for calculating distances easily. Particularly: 
Let $x$ and $y$ be two independent q-states, then
\begin{eqnarray}\label{eq:Distance}
&& d(x,y) \;=\; 2 \: q \: (1-q)\\
&& d(x,x*y) \;=\; 1-q.
\end{eqnarray}

\begin{figure}[h]
{\center
\includegraphics[width=2.8in]{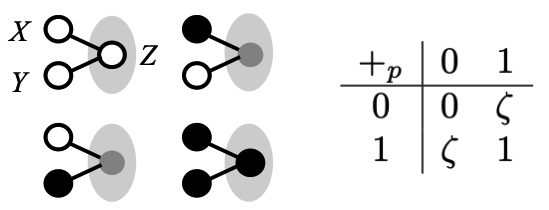} 
\caption{Bundling patterns by adding activities: neuron $Z$ is exposed to a variety of other neurons, the corresponding 'heat bath' is displayed by the greyish ellipse. Moreover, $\zeta \in \{0,+1\}$ is random with ${\mathbb P}[\zeta=1] = p$. }\label{fig:addition}
}
\end{figure}

\paragraph{\bf Bundling by addition}
Bundling mimics stochastic addition of activity patterns as shown in Fig \ref{fig:addition}. Assume that two neurons $X$ and $Y$ converge on a third neuron $Z$, which is exposed to activating input also from other neurons, the 'heat bath' being indicated by the greyish ellipse around neuron $Z$. This setting makes addition a probabilistic. If both neurons are inactive, i.e. $x=y=0$, neuron $Z$ will also be, $z=0$, while if both are active, $Z$ will be active, i.e. $1+1 = 1$. If only $X$ or $Y$ is active, the activation of $Z$ is probabilistic and depends on some threshold: If the activation threshold is low, $Z$ is likely to be active, while if the activation threshold is high, $Z$ will be inactive. Formally, the threshold $p$ equals the probability that $z$ is inactive, i.e. $p = \mathbb{P}[z=0]$. Correspondingly, addition is defined as a stochastic mixture of $AND$ and $OR$. Obviously, if $p=1$, addition is component-wise $AND.$, while for $p=0$, addition is component-wise $OR.$.This bundling operation therefore generalises the one used in Binary Sparse Distributed Codes \cite{rachkovskij2001representation}. Note that in the Binary Spatter Code \cite{kanerva1994spatter} addition is realised by a deterministic normalisation procedure.

Note that for $p=0$, the bundling operation equals the component-wise $AND.$, while in case of $p=1$, bundling equals component-wise $OR.$. Both operations are associative. 
It moreover follows hat bundling is contracting, i.e. for two independent q-states $x$ and $y$
\begin{equation}\label{eq:DistanceAddition}
d(x,x \oplus_p y) = 2q(1-q)(1-p) \leq d(x,y), 
\end{equation} 
where equality holds for $p=0$.


This completes the definition of the algebra used to calculating with cognitive states.
\begin{equation}\label{eq:algebra}
\Big( \mathbb{X}^N_q,\oplus_p,\ast \Big) 
\end{equation}

For consistence it is necessary to show that the two operations play the role or addition and multiplication. Their defined properties are, roughly speaking, that addition preserves similarity, while multiplication creates similarity \cite{plate1997common}. In the following important properties are summarised: States are q-states, i.e. $x,y \in \mathbb{X}^N_q$, where $N$ is large and $0 < q  < 1$. Particularly $1 \in \mathbb{X}^N$ is the vector with all entries being $1$. Furthermore $0 \leq p \leq 1$. Under these conditions the following holds: 

{\it 
\begin{enumerate}  
\item Addition and multiplication are commutative, while each state is its own neutral element in addition as well as its own multiplicative inverse, i.e. 
\begin{equation}
x \oplus_p x = x \qquad x \ast x = 1,
\end{equation}

\item Multiplication approximately distributes over addition, i.e. for large $N$ 
\begin{equation}
x*(y \oplus_p z) \approx x*y \oplus_p x*z
\end{equation}
\item For sparse states, $q < \frac{1}{2}$, addition decreases distance, while multiplication increases distance.
\begin{equation}
D(x,x \oplus_p y) < D(x,y) < D(x, x\ast y).
\end{equation}
\item The more sparse states are, the more distinct are the states resulting from addition and multiplication.
\begin{equation}
d(x \oplus_p y, x*y)\geq 2 p (1-q)
\end{equation}
\item Addition $\oplus_p$ is not associative, unless $p=0,1$.
\end{enumerate}
}

In the following we therefore restrict the parameter range to $ 0 < q <\frac{1}{2}$ and $0 < p < 1$. Later considerations will impose additional constraints on the parameter setting. The proof of the above statements can be found in the Appendix. 

\section{Implied properties of bundles}\label{sec:bundles}
Having defined addition on the state-space, we can consider bundles of states and their properties implied by the particular addition of respective states. We will only consider the non-associative case, i.e. $p \not= 0,1$. If the bundling operation is non-associative, a sequence of states gives rise to two bundles, the bundle $\bf L$ originating from left-associative addition and the bundle $\bf R$ from right-associative addition. For concreteness consider the sequence of states $(a,b,c,d,e)$. The two corresponding bundles are
\begin{eqnarray}
	{\bf L} &=& \big(((\eta \oplus_p a) \oplus_p b ) \oplus_p c )\oplus_p d\big) \oplus_p e \\
	{\bf R} &=& \eta\oplus_p \big(a\oplus_p ( b \oplus_p  ( c \oplus_p ( d \oplus_p e )))\big) 
\end{eqnarray}
where each state is constructed from an initial state $\eta$.

\subsection{The distance profiles}\label{seq:distances}
\begin{figure}[h]
{\center
\includegraphics[width=2.2in]{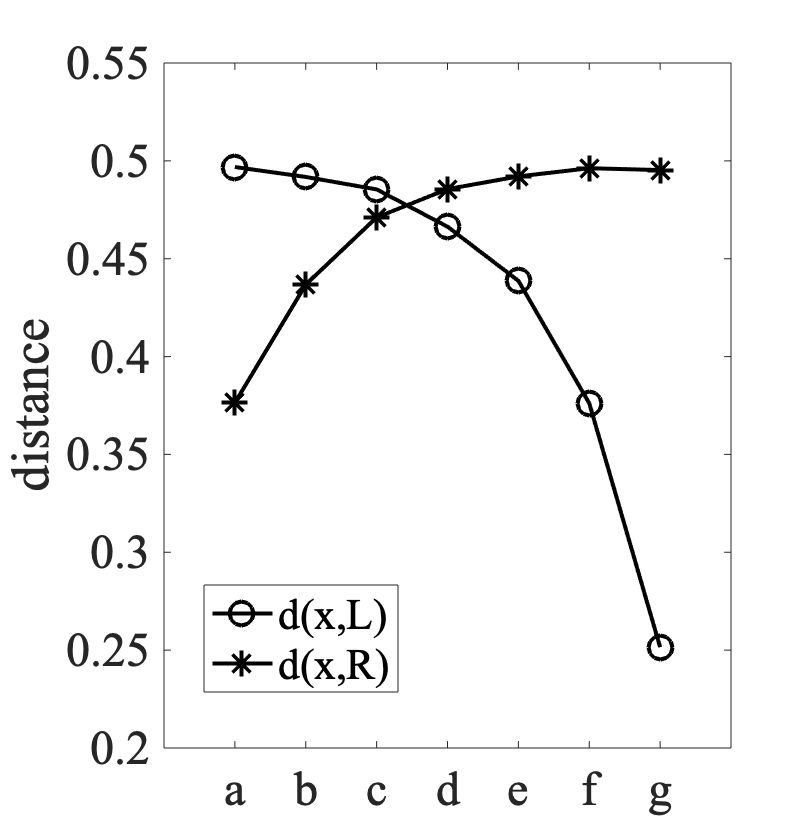}
\caption{{\bf Distance gradients} implied by the non-associativity of bundling, $p = \frac{1}{2}$, while states are dense. $\bf L$ has smallest distances to the most recent items, while $\bf R$ is closest to the early list items. Distance gradients directly translate into recency and primacy gradients, respectively, see below
}\label{fig:DistancesGradients}}
\end{figure}

In case of non-associativity, one expects that both states code sequential information. They in fact do as seen by their respective distance profiles, see Fig. \ref{fig:DistancesGradients}. For any state $x$ in the sequence $(a,b,\hdots)$, $d(x,{\bf L})$ and $d(x,{\bf R})$ are the distances of $x$ from the respective bundle. Moving along the sequence $(a,b,\hdots)$, distances increase for the $\bf R$ bundle, while they decrease for the $\bf L$ bundle. $\bf L$ has smallest distances to the most recent items, while $\bf R$ is closest to the early list items. That is, the information stored in $\bf L$ is most similar to the most recent items, while information stored in $\bf R$ is closest to the earliest ones. If bundling were associative, for symmetry reasons profiles are identical and flat. The reason that the profiles are not constant is that in each summation step, noise is injected into the bundle, which during the iterative construction the memory state accumulates. Therefore distances vary monotonously along serial position, while the states ${\bf R}$ and ${\bf L}$ become independent, i.e. quasi-orthogonal, when the list length increases.   

\begin{figure}[h]
{\center
		\includegraphics[height=2.2in]{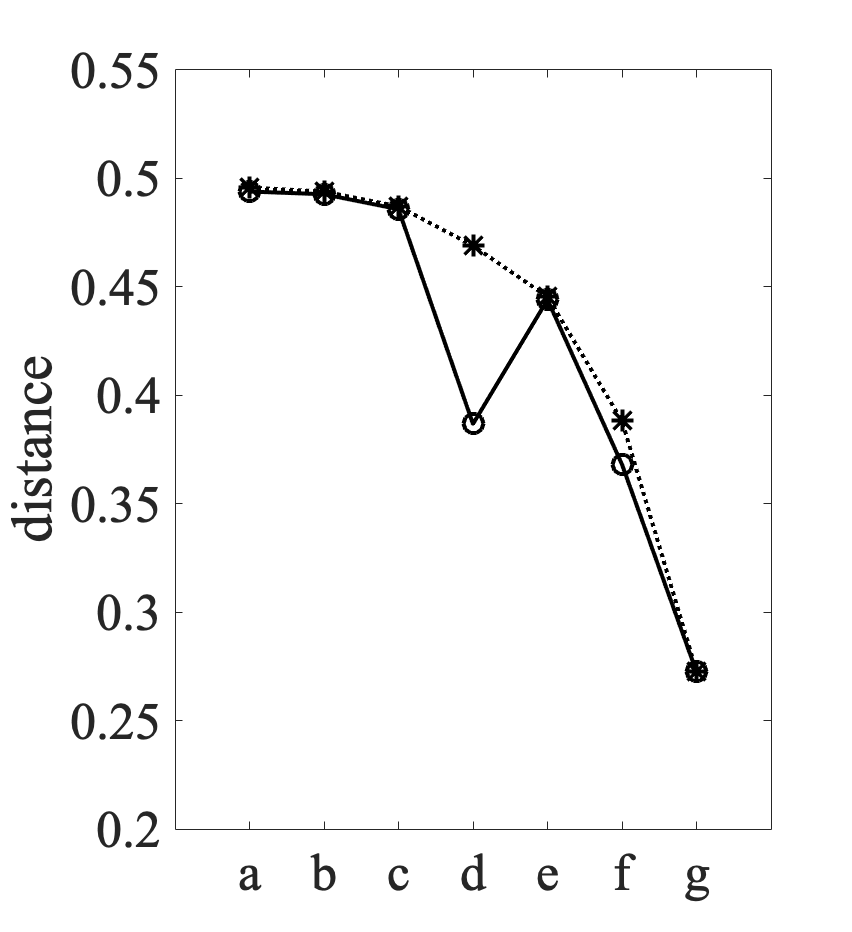}
		\caption{Distance profiles of the $\bf L$-state ($\star$) and the $\tilde{\bf L}$ state ($\circ$). The similarity of $d$ to $f$ results in a clear peak at $d$.}\label{fig:SimCurves}
		}
\end{figure}

While the distance profile is sensitive to sequential ordering, it is also sensitive to similarities between items. In Fig. \ref{fig:SimCurves} the distance profiles are shown for the bundle $\bf L$ resulting from the sequence of independent items $(a,\hdots,g)$ and the bundle $\tilde{\bf L}$, which results from the same list, where state $d$ is similar to state $f$ shows a peak for item $d$, which is similar to item $f$. It thus reacts to both, item information about similarity as well as to order information. 

\begin{figure}[h]
{\center
\includegraphics[height=2.2in]{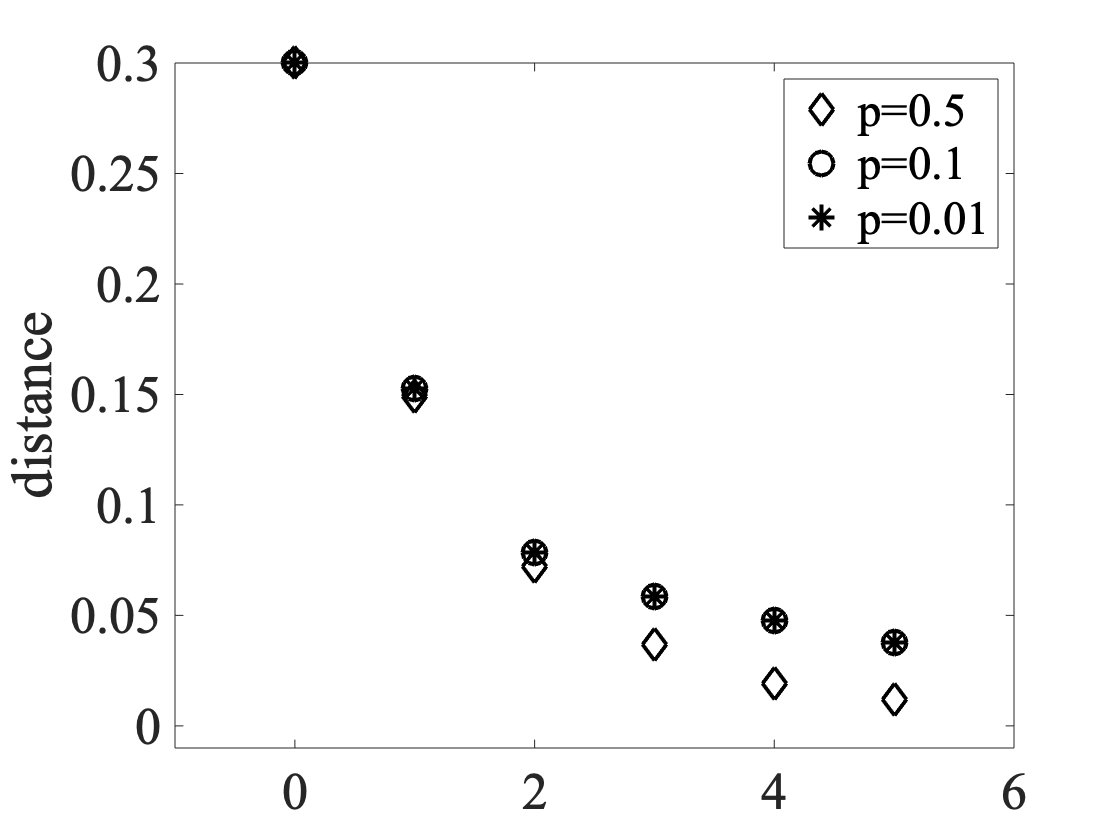} 
\caption{{\bf Convergence:} The distance is $d(y,{\bf L}_m)$ and measures the convergence of the $\bf L$-state as defined in eq. \ref{eq:conv} towards $y$ for various values of $p$. Convergence happens fast 
}\label{fig:BNN_distance}}
\end{figure}

The contracting property of bundling can be used to 'navigate' a state $x$ towards another, $y$ say. Consider the particular sequence $(y,\hdots,y)$ where $m$ is the number of states $y$ driven the initial state $x$. Then 
\begin{eqnarray}\label{eq:conv}
{\bf R}_m &=& x \oplus_p (y \oplus_p (y \oplus_p \hdots \\
 {\bf L}_m &:=&\big((x \oplus_p y) \oplus_p y )\hdots \big)\oplus_p y .
\end{eqnarray}
Then ${\bf R}_m = x \oplus_p y$ for any $m>0$, while $ {\bf L}_m$ converges to $y$ exponentially, see Fig. \ref{fig:BNN_distance}
Recalling that the larger $p$, the larger the noise level in addition, it is seen that noise speeds up the convergence towards the target $y$. 

\subsection{Bundles are sparse}\label{sec:sparseness}
In artificial settings such as experiments, list length is finite or even restricted to very low numbers such as $5$ or $8$. In real world, the cognitive agent is subject to a continuous inflow of information, i.e. the input sequence has arbitrary length. The question therefore is: Can bundling an arbitrary number of informative states result in a homogeneous memory state? The answer is 'No, provided that addition is not-associative!', i.e. $p \not= 0,1$. 
Moreover, the average activity is controlled by the bundling operator, since $p$ equals the probability that a component is active ($1$). Hence, decreasing $p$ will decrease average activity in the bundle. Sparseness of bundles can thus be fine-tuned by this parameter. The importance of sparseness is nowadays well established and documented for a huge number of neuronal systems \cite{olshausen2004sparse}. From a cognitive point of view it is desirable, to find a degree of sparseness which realises both: it maximises capacity, while it minimises energy consumption, \cite{foldiak2003sparse,palm2013neural}. For this reason, fine tuning of sparseness is important. 

The main observation about successive bundling and evolving average activity is the following:\\

{\bf Bundling creates sparseness:}
{\it Let $\lambda_k$ be the list of $k$ independent q-states representing information items, i.e. $\lambda_k = \left( x_1, x_2, \hdots, x_k\right)$, where $k>0$.  
Bundling of states is done by addition $\oplus_p$ with parameter $0 < p < 1$. Let ${\bf M}_k$ be the bundle representing the list $\lambda_k $ and $Q({\bf M}_k)$ its average activity. If the list length increases, the average activity converges exponentially to
\begin{equation}\label{eq:Q}
Q({\bf M}_k) \to \frac{p \: q}{1-p(1-q)-q(1-p)}, \qquad k \to \infty
\end{equation}
Particularly, the memory state is non homogeneous for arbitrary many bundled items.
If $p = \frac{1}{2}$, the average activity is independent of list length, particularly $Q({\bf M}_k)=q$ for all $k$. Particularly, if q-states are dense, $q=\frac{1}{2}$,  
\begin{equation}\label{eq:Q2}
Q({\bf M}_k) = p - \left( p-\frac{1}{2}\right) 2^{1-k}, \qquad k>0.
\end{equation} 
}

\begin{figure}[h]
{\center
\includegraphics[height=2.2in]{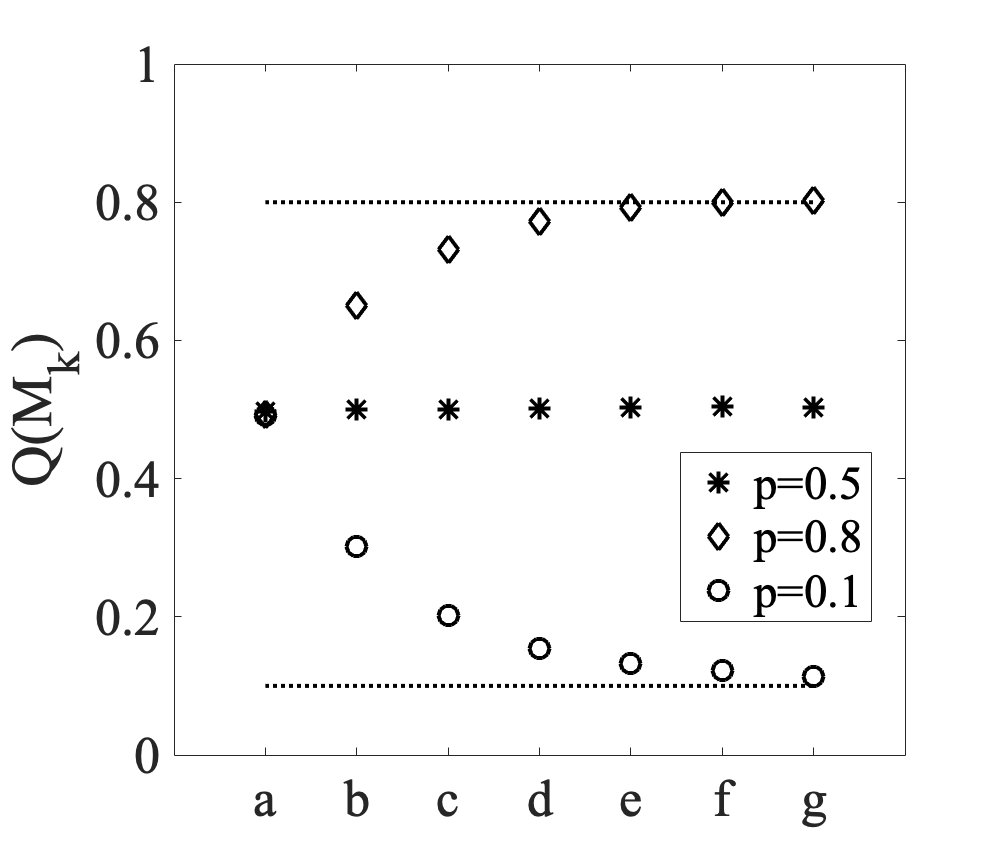} 
\caption{{\bf Bundles are sparse.} Given a list $(A,B,\hdots,G)$, all items are represented by dense states $(a,b,\hdots,g)$. The bundle ${\bf M}_k$ is the superposition of the first $k$ states, governed by $\oplus_p$. The curve corresponds to different $p$ values and follows eq. \ref{eq:Q2}. \label{fig:SparseMemory}}
}
\end{figure} 

The argument for eq.\ref{eq:Q} is the following: Let $P(n)$ denote the probability that the sum of $n$ components is $1$, then 
{
\begin{eqnarray*}
P(1) &=& P[x_1=1] = q\\
P(2) &=& P[x_1 + x_2 = 1] = (1-2p)q^2 + 2pq;\\
&\vdots&\\
P(n) &=& q P(n-1) + \hdots \\
&& + p \big[(1-q)P(n-1) + q (1-P(n-1)]\big)
\end{eqnarray*}
}
whose generating function is 
\begin{equation}
G(s) := \frac{qs + \frac{qps^2}{1-s}}{1-p(1-q)-q(1-p)}
\end{equation}
Consequently $Q({\bf M}_k) = \frac{1}{k!} G^{(k)}(0)$ from which eq. \ref{eq:Q} follows. 
Moreover, 
$P(n)|_{p=\frac{1}{2}} = q$ for all $n$, while $P(n)|_{q=\frac{1}{2}} = p - \left( p-\frac{1}{2} 2^{1-n}\right) \to p$ for $n \to \infty$. 

The addition parameter $p$ plays a major role in fine tuning the sparseness of the resulting memory state. Since sparse states are preferable from an energetic view, plausible parameters are 
\begin{equation}\label{eq:range}
0 < p,q \leq \frac{1}{2}.
\end{equation}
In that range, bundling of an arbitrary number of dense q-states results in a non-homogeneous, sparse memory state.

\section{Conclusion}
This note refers to the model about bundling information states originally proposed in the context of human working memory, see \cite{reimann2021algebra}. The bundling operation, i.e. addition of states representing formation items, is inspired by the binary stochastic summation of neuronal activities. Being stochastic addition injects noise into the system. The noise level is governed by a parameter $p$, where for $p=0,1$ the bundling operations are component-wise $AND.$ and $OR.$, respectively. In these two cases, bundling is associative, so that the sum is independent of the ordering of its components. Otherwise bundling is non-associative, so that the bundle inherits information about ordering. In that case, the information state resulting from bundling items encodes item information as well as sequential information. If $p=0,1$, the bundle resulting from an infinite number of items is homogeneous, i.e. identical to the zero-vector or the unity vector. This bundling catastrophe does not occur in the non-associative case. 

When it comes to memory, representational capacity and memory capacity have to be jointly maximised under the constraint of limited 'energy', e.g. average neuronal activity. For example, dense states have a high representational capacity but a low memory capacity. The elementary mechanism for bundling items is shown to create sparseness, while the corresponding parameter allows to fine tune the degree of average activity. The required fine tuning of sparseness can be achieved by varying the 'noise parameter' $p$. Recall that this parameter can be regarded as being related to an activation threshold for neurons. Obviously, increasing this threshold decreases the probability for activity and hence increases the level of sparseness. 

Memory states, i.e. bundles constructed from a sequence of information items in the parameter range given in eq. \ref{eq:range}, represent both: item information as well as sequential information, and are sparse vectors. 

That bundle should not be regarded as a storage medium, since only few items can be retrieved from it with sufficient accuracy, either the most primary items or the most recent ones.  Being sensitive to items' similarities inside that 'temporal' window, its role appears to be more likely to that of a filter in both, in the 'item domain' as well as in the 'temporal domain'. Given that real world cognitive systems, biological as well as technical, are subject to a continuous inflow of information, this filter allows to navigate that stream of high dimensional information by weighting similarities in item information as well as their temporal recency. 


\end{document}